\documentclass[lettersize,journal]{IEEEtran} 
\usepackage[utf8]{inputenc}
\usepackage{tikz} 
\usepackage{subfigure}
\usepackage{mathtools}          
\usepackage{acro}               
\usepackage{comment}
\usepackage{amsmath}
\usepackage{amssymb}
\usepackage{xcolor}
\usepackage{listings}
\usepackage[colorlinks = true,
            linkcolor = blue,
            urlcolor  = blue,
            citecolor = blue,
            anchorcolor = blue]{hyperref}
\usepackage[font=small]{caption}
\usepackage{url}

\title{PSGSL: A Probabilistic Framework Integrating Semantic Scene Understanding and Gas Sensing for Gas Source Localization}
\author{Pepe Ojeda, Javier Monroy and Javier Gonzalez-Jimenez}

\twocolumn

\begin{document}

\maketitle

\begin{abstract}
    Semantic scene understanding allows a robotic agent to reason about problems in complex ways, using information from multiple and varied sensors to make deductions about a particular matter. As a result, this form of intelligent robotics is capable of performing more complex tasks and achieving more precise results than simpler approaches based on single data sources. However, these improved capabilities come at the cost of higher complexity, both computational and in terms of design. Due to the increased design complexity, formal approaches for exploiting semantic understanding become necessary.
            
    We present here a probabilistic formulation for integrating semantic knowledge into the process of gas source localization (GSL). The problem of GSL poses many unsolved challenges, and proposed solutions need to contend with the constraining limitations of sensing hardware. By exploiting semantic scene understanding, we can leverage other sources of information, such as vision, to improve the estimation of the source location. We show how our formulation can be applied to pre-existing GSL algorithms and the effect that including semantic data has on the produced estimations of the location of the source.
\end{abstract}

\section{Introduction}
The problem of finding the source of a gas emission with an autonomous mobile robot (\textit{GSL}, for gas source localization) is a challenging one. Because of the current state of sensing technology, the sensory information available to the robot is usually limited to a single-point gas concentration measurement~\cite{Karakaya2020}. Due to the complexity of the underlying problem of describing the gas dispersion through the environment, this kind of sensing does not provide enough information to correctly estimate the location of the emitting source, especially in realistic situations where the gas is not merely being diffused, but also advected by a complex airflow. Moreover, depending on the desired selectivity of the sensor, a recovery time might be needed after each gas measurement~\cite{gmonroy2013mox,Martinez2019}, further complicating the search task. 

Because of these limitations that gas sensors impose, the concept of sensor fusion is particularly relevant to the problem of GSL. Sensors of a different type can make up for the relative scarcity of information available to the robot, allowing for more efficient search methods to be devised. To make use of the data that can be obtained from other types of sensors, however, we must define a connection between the information that those sensors provide and the release and dispersal of gases.

In this work, we consider the subject of semantic scene understanding as a basis for this sensor fusion. By defining a set of relevant concepts and the relationships between them, it is possible to use information that is not directly related to gas sensing in the process of locating the source of a gas release. In particular, we focus on the idea of leveraging semantic maps, which are representations of the environment that contain information about the types of objects or rooms that certain regions correspond to. With the help of an ontology that defines information such as which types of objects can release certain smells, we can apply visual sensing or pre-existing information about the environment to quickly estimate the source location and achieve higher precision in declaring the final source position.

Our proposal does not consist of a specific GSL method that includes semantic information into its estimations of the source location, but instead, an abstract probabilistic framework that formally defines how semantic information can be fused with any GSL algorithm. By keeping the details of the olfaction-based estimations of the source location abstracted away, the formulation we present here achieves a greater level of flexibility that makes it more widely applicable. 

Our contributions are:

\begin{itemize}
    \item A formally defined probabilistic formulation that shows how olfaction-based and semantics-based estimations of the source location can be integrated, regardless of the specifics of each of the two methods (section~\ref{sec:framework}).
    \item A method that leverages a per-cell semantic map and an ontology to estimate the source location, independent of olfactory information (section~\ref{sec:semantics}). We also show several ways in which this method can be extended in section~\ref{sec:extensions}.
    \item We discuss the calculation of expected information gain when using the previously presented semantics method, for the design of efficient navigation strategies.
    \item A set of experiments that show the effect of including semantic information in the estimations of the source location produced by a state-of-the-art GSL algorithm.
\end{itemize}

\section{Related Work}
Probabilistic algorithms are one of the most popular approaches to performing gas source localization in recent years~\cite{francisGasSourceLocalization2022a,wangExplorationEnhancedSearchAlgorithm2024,jinEfficientGasLeak2023}. Because of their greater robustness when dealing with imperfect sensory information and their ability to make predictions about the source location and the state of the environment, these algorithms are a more sophisticate and interesting approach than the pure navigation bio-inspired strategies that were favored in previous decades~\cite{kowadloRobotOdorLocalization2008}.

An additional benefit of using formally defined probability distributions to maintain and update a belief about the source location is that it becomes possible to use these same tools of probability theory to extend the algorithms with additional information. 

It is interesting to note that the idea of including visual object detection into a GSL algorithm was first explored in the mid-2000s~\cite{Ishida2006,Loutfi2004}. Some algorithms~\cite{Ishida2006} proposed to use the detection of objects to alter the behavior of the robot, from the state-action perspective. However, these algorithms consisted mostly on navigation strategies, and only used the detected objects to guide the movements of the robot, rather than to identify and declare gas sources. More recent proposals~\cite{Monroy2018Semantics} have included the object detection into a probabilistic framework that exploits semantic knowledge, but were oriented to provide an optimized path to visit all known candidate sources, rather than to explore the environment and dynamically update a belief about the source location. 

The approach we present here to leverage visual observations as part of the source localization process relies on scene understanding. This is a complex process, and several types of techniques are relevant to it:

\begin{itemize}
    \item \textbf{Object recognition}: For years, Deep Convolutional Neural Networks~\cite{zhao2019} have been the gold-standard tool for performing object recognition and segmentation. In recent times, transformers have shown to also be a viable option for these tasks~\cite{LiTransformers2023}, similarly relying on machine learning.
    \item \textbf{Scene representation}: Semantic maps can be based on different types of representations, including sparse instance collections~\cite{wang2019_multi} and dense, per-voxel information~\cite{grinvald2019_volumetric,mascaro2022_volumetric}. Of particular interest to our proposal here is the idea of performing the semantic mapping in a probabilistic manner~\cite{voxeland}, since it allows us to rigorously integrate this information into our framework and formally handle uncertainty.
    \item \textbf{Semantic understanding}: The subject of semantic understanding in artificial intelligence has traditionally been tackled through the use of hand-crafted ontologies~\cite{Monroy2018Semantics}. An ontology is a representation of the concepts that are relevant to a specific field or problem and the relations that may exist between them~\cite{liu2023survey}. 
    
    In recent times, the advent of Large Language Models (LLMs) has created a promising alternative~\cite{ZengeyaLearningOntology}, where the relations between the different concepts are learned and implicitly represented by the embedding mechanism. A particularly relevant technique related to this is Large Vision-Language models (LVLMs)~\cite{liu2024visual}, which can encompass the object recognition task as well.  
\end{itemize}

Our formulation assumes that it is possible to obtain a dense probabilistic map of objects, but makes no assumptions about the tools utilized to generate it. Similarly, and although we will use the term ``ontology'' to denote the semantic model that relates object types to gases, we do not require that a hand-crafted ontology is used, and the formulation is equally applicable to machine-learning-based semantic models.


\section{Probabilistic Formulation}
\label{sec:framework}
There are numerous methods~\cite{francisGasSourceLocalization2022a,jinEfficientGasLeak2023,Hutchinson2017,Ojeda2021RAL,OjedaPMFS} which can estimate a probability distribution for the source location based on olfactory methods, which we can denote $p(s | g)$. Our intuition tells us that it would also be possible to design a method that produces a similar estimation based on semantic data and information about the environment, $p(s | z)$. However, combining both of these sources of information is not trivial. 

One possibility is to design a novel GSL method which integrates them into a single model, defining a specific calculation of $p(s| g, z)$. A more flexible approach, on the other hand, would be to express the conditional probability distribution of the source location given all types of measurements as a function of two separate terms, where one of the terms depends only on the olfactory data and one depends only on semantic data:

\begin{equation}
\label{eq:decomposition-objective}
    p(s_c | g, z) \stackrel{?}{=} f\big(p(s_c | g), p(s_c | z)\big)
\end{equation}

With such a decomposition, it would be possible to combine any form of pure-olfaction source localization (including pre-existing algorithms) with any form of semantic source localization, rather than it being necessary to design specific methods that simultaneously depend on both.

\begin{figure*}[t]
    \centering
    \includegraphics[width=\linewidth]{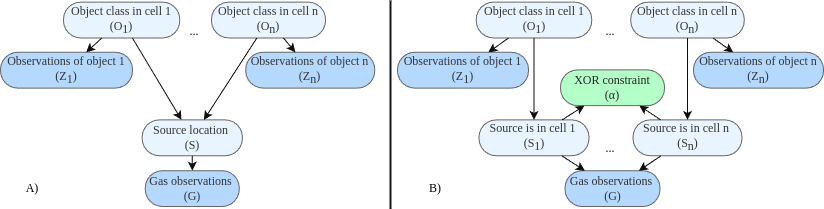}
    \caption{Bayesian networks modeling the conditional dependencies of the random variables involved in the problem. 
    Network A models the source location through a single variable $S$ with a discrete domain, while network B uses a set of binary variables with an added XOR constraint to achieve desirable d-separation properties.}
    \label{fig:Bayes_net}
\end{figure*}

\subsection{Model}
\label{sec:prob-model}
We start by considering a division of the environment into a grid of $n$ equally-sized cells. Each cell $c$ contains an object $O_c$ that can be of any of considered object categories. We want to estimate each cell's probability of containing the gas source, and if we assume a single gas source, the most intuitive way to model this is through a single variable $S \in \{c_1,\dots c_n\}$. The estimation of this probability distribution will be derived from the sensor measurements, and we make a distinction between two main types of sensory input: olfaction data and object categorization data. 

Olfaction observations (which we will denote $G$) may consist of different information depending on which source localization method we wish to employ. Essentially, $G$ represents all the information necessary for the GSL method to provide $p(S=s | G=g)$ as its output. This means $G$ can be single-point gas concentration readings, binarized \textit{hits}, or even maps of the distribution of the gas in the environment. Similarly, it can optionally include airflow information.

Object categorization observations ($\boldsymbol{Z}$) will often consist of images (both color and depth) gathered by the robot as it navigates the environment. However, this can also include any other information that is relevant for updating the belief about the object class of a given cell -- any observation such that we can provide a value for $p(O_c | \boldsymbol{z})$. For example, if a room categorization map is supplied ahead of the search (which would modify the belief of the object categories based on which type of room the cell is in), this could be considered to be an observation in $\boldsymbol{Z}$ as well. It should also be noted that the set $\boldsymbol{Z}$ can be divided into $n$ disjoint sets $\boldsymbol{Z} = \{\boldsymbol{Z}_1,\dots \boldsymbol{Z}_n\}$ where $\boldsymbol{Z}_i$ corresponds to the observations of $O_i$.  This distinction is important for the proof in the following section.

The probabilistic model described so far can be represented by the Bayesian network in Fig.~\ref{fig:Bayes_net}A. However, this model does not correctly represent all the conditional independence relationships between the variables --particularly, the independence of $O_i$ and $O_j$ given the value of the source location variable $S$ is known. It is thus convenient for our purposes to modify the model as shown in Fig.~\ref{fig:Bayes_net}B. In this case, $S$ --which had $\{c_1,\dots c_n\}$ as its possible values-- has been decomposed into $n$ binary variables $\boldsymbol{S} = \{S_1,\dots S_n\}$, where $S_i$ represents whether cell $i$ is a source of gas or not. In this network, $O_i$ and $O_j$ are d-separable given the value of $S_i$ or $S_j$, and thus conditionally independent.  

To make this model equivalent to the previous one, however, it is necessary to enforce that a single source exists. This can be done by adding a constraint variable $\alpha$ (see~\cite{jensenBayesianNetworks2007}, chapter 3) such that $p(\alpha | s) = 1$ for any configuration of $S$ that has a single cell set to true, and $p(\alpha | s) = 0$ otherwise (XOR constraint). The constraint is enforced by instantiating $\alpha =  1$.

\subsection{Separating the Semantic and Olfactory Estimation Processes}
We define $\boldsymbol{O} = \{O_1,\dots O_n\}$ to be the vector set of all the random variables for the object class of each cell, and $o_M$ is one specific value for that map. Though unusual, we are using a capital letter M for an index to make explicit that we are referring to the whole map.   
The probability that $S$ takes a specific value $s$ given the gas observations ($g$), the object observations ($z$) and the X-OR constraint ($\alpha$) can be calculated by marginalizing out the objects map: 
\begin{equation}
    p(s | g, z, \alpha) = \sum_{M} p(s | g, z, \alpha, o_M) \cdot p(o_M | g, z, \alpha)
\end{equation}

Then applying Bayes' Theorem to the second term:

\begin{equation}
    p(s | g, z, \alpha) = \sum_{M} p(s | g, z, \alpha, o_M) \cdot \frac{p(g | z, \alpha, o_M) \cdot p(o_M|z, \alpha)}{p(g | z, \alpha)} 
\end{equation}

The term $p(g | z, \alpha)$ can be factored out of the sum, and since it is constant for all values of $s$, it can be omitted from the calculation by including it in the normalization factor. We can also remove some conditionally independent variables from the expression as per the previously defined Bayesian network:

\begin{equation}
    p(s | g, z, \alpha) \propto  \sum_{M} p(s | g, \alpha, o_M) \cdot p(g |\alpha, o_M) \cdot p(o_M|z, \alpha)
\end{equation}

Applying Bayes' theorem now to the first term:

\begin{equation}
\begin{split}
    p(s | g, z, \alpha) \propto \sum_{M} \frac{p(g | s, \alpha, o_M) \cdot p(s| \alpha, o_M)}{p(g | \alpha, o_M)} \cdot  p(g |\alpha, o_M) \cdot p(o_M|z, \alpha)
\end{split}
\end{equation}

At this point, the denominator of the first term cancels out with the second term, and several conditionally independent variables can be removed:

\begin{equation}
\begin{split}
    p(s | g, z, \alpha) &\propto \sum_{M} p(g | s) \cdot p(s| \alpha, o_M) \cdot p(o_M|z, \alpha) \\
    &\propto  p(g | s) \cdot \sum_M p(s| \alpha, o_M) \cdot p(o_M|z, \alpha) \\
    &\propto  \frac{p(s | g) \cdot p(g)}{p(s)} \cdot \sum_M p(s| \alpha, o_M) \cdot p(o_M|z, \alpha) \\
\end{split}
\end{equation}

Assuming a uniform a priori probability for the source location, the prior terms $p(s)$ and $p(g)$ can get included in the normalization factor:

\begin{equation}
    p(s | g, z, \alpha) \propto p(s | g) \cdot \sum_M p(s| \alpha, o_M) \cdot p(o_M|z, \alpha) 
\end{equation}

At this point, since $S$ is independent of $Z$ given $o_M$ as per our previously defined Bayesian network, we can add $z$ as a condition to $p(s| \alpha, o_M) = p(s|z, \alpha, o_M)$. This allows us to apply the law of total probability to this expression so the sum over $O$ becomes equal to $p(s | z, \alpha)$:

\begin{equation}
\begin{split}
    p(s | g, z, \alpha) &\propto p(s | g) \cdot \sum_M p(s| z, \alpha, o_M) \cdot p(o_M|z, \alpha) \\
    &\propto p(s | g) \cdot p(s | z, \alpha)
\end{split}
\end{equation}

This expression is almost what we are looking for, as we have a clean separation between $p(s | g)$, which can be provided by any pure-olfaction GSL method, and $p(s|z, \alpha)$, which does not depend on the olfactory data. However, we still need to compute the second term, which has a dependency on the constraint variable. With the conditional independence of $Z$ and $\alpha$ given $S$ defined by our Bayesian network, we can show that:

\begin{equation}
\begin{split}
    p(s | z, \alpha) &= \frac{p(z,\alpha | s) \cdot p(s)}{p(z,\alpha)} \\
    &\propto  p(z,\alpha | s) \\
    &\propto  p(z | s) \cdot p(\alpha | s) \\
\end{split}
\end{equation}

The value of $p(\alpha | s)$ is given by the definition of the variable as a constraint. Specifically, $p(\alpha | s) = 1$ for valid configurations of $S$ and $p(\alpha | s) = 0$ otherwise. Thus, for any value of $S$ (any value with a number of sources different than one), the value of $p(s | g, z, \alpha)$ is trivially 0. Any value $s$ that satisfies the constraint will be of the form $s_c$, meaning it specifies a single source in cell $c$. For these values, we can apply Bayes' Theorem again to reach a more interpretable expression:

\begin{equation}
\begin{split}
    p(s_c | g, z, \alpha) &\propto p(s_c | g) \cdot p(z | s_c) \\
    &\propto p(s_c | g) \cdot \frac{p(s_c | z) \cdot p(z)}{p(s_c)} \\    
    &\propto p(s_c | g) \cdot p(s_c | z) \\    
\end{split}
\end{equation}

Thus, we finally arrive at the desired conclusion that we can calculate the probability distribution of the source location given each type of sensory information independently, and combine them through simple multiplication and normalization:

\begin{equation}
\label{eq:combined-conditionals}
    p(s_c | g, z, \alpha) \propto p(s_c | g) \cdot p(s_c | z)
\end{equation}

\vspace{\baselineskip}
\section{Source Localization from Semantic Data}
\label{sec:semantics}
In this section we will discuss the probabilistic estimation of the source location based only on semantic information as per the proposed framework, and show that a computationally light method can be employed for this calculation.

\subsection{Semantic Knowledge and Ontology}
\label{sec:ontology}
In order to relate the concept of a gas source to a map of objects, we must set which types of objects can emit a certain gas --or, in general, a smell--, and with what probability.

We rely on an ontology to represent this relationship. Specifically, for a given smell, we define the probabilities of it being emitted by each of the different object classes we consider. The specific values for these probabilities can be obtained by analyzing data on known releases with a frequentist approach, or simply be based on the judgment of an expert. Formally, this probability can be expressed as $p(o_{ci}|s_c)$.

As a clarification example, consider a case in which the robot is looking for the source of smoke in an industrial kitchen, and the semantic map can contain objects of the following classes: \textit{oven}, \textit{microwave}, \textit{refrigerator}, \textit{countertop}. The simplest version of an ontology to cover this case is simply a probability distribution over these object classes for the source cell:

\vspace{1cm}
{
\small
\vspace{\baselineskip}
$p(o_{ci} | s_c)$:
\begin{lstlisting}[]
    - oven: 0.4
    - microwave: 0.25
    - refrigerator: 0.3
    - countertop: 0.05
\end{lstlisting}
}

A more complex ontology that considers the relations between the objects themselves, such as whether multiple object categories are subcategories of a common abstract concept, could be used as well, as long as it is possible to extract from it a probability value for $p(o_{ci} | s_c)$. Similarly, it is not necessary for the ontology to assume one specific gas type. Instead, the ontology can define multiple probability distributions of the form $p(o_{ci} | s_c, \gamma)$, where $\gamma$ may be any of the considered gas types. Section~\ref{sec:gas-classification} describes how such an ontology is included in the formulation.

The ontology could optionally also include additional semantic information. For example, it could specify the a priori probability of an object being of each of the considered classes, or the conditional probability of these object classes given a particular room type (\textit{e.g.} kitchen, bathroom, corridor, \textit{etc.}). In the absence of this data, one could simply assume an equiprobable prior distribution for the object classes and omit any room classification information from the set of semantic evidence $Z$.

\subsection{Derivation of the Distribution  $p(s_c | z)$} 
The value that we are interested in is $p(s_c | z)$. We cannot directly calculate this, as there is no trivial way to connect the source location ($S$) to the observations ($Z$). Instead, the two variables are connected through the object categories ($O$). We could, as in the previous section, marginalize over the set of possible object maps to exploit conditional independence of $S$ and $Z$, such that:

\begin{equation}
\begin{split}
    p(s_c|z) &= \sum_M p(s_c| z, o_M) \cdot p(o_M | z) \\
    &= \sum_M p(s_c| o_M) \cdot p(o_M | z)
\end{split}
\end{equation}

However, this expression is computationally intractable, as the number of possible object maps for $n$ cells and $k$ object categories is $k^n$. Instead, we can use the distinct subsets of observations introduced in Section~\ref{sec:prob-model}, where we defined that $\boldsymbol{Z} = \{\boldsymbol{Z}_1, \dots \boldsymbol{Z}_n\}$ such that observations $\boldsymbol{Z}_i$ are only related to $O_i$. When working with images, where a single image will most likely contain observations of multiple cells, one could conceptualize this decomposition as treating each region of the image as a separate observation, such that one observation corresponds to one cell only.

We can exploit the conditional independence of the subsets of $\boldsymbol{Z}$ given $\boldsymbol{S}$ by separating $\boldsymbol{Z}$ into multiple disjoint sets depending on which cell is observed by the measurement. We use $z_n$ to denote the set of all observations of cell $n$ and $z_{\neg n}$ to denote all the remaining observations: 
\begin{equation}
\begin{split}
    p(s_c|z) &\propto p(z|s_c) \\
    & \propto p(z_n, z_{\neg n} | s_c) \\
    & \propto p(z_n | s_c) \cdot p(z_{\neg n} | s_c) \\
\end{split}
\end{equation}

We can repeat this process for each cell, arriving at the following expression:

\begin{equation}
\begin{split}
    p(s_c|z) & \propto \prod_{n} p(z_n | s_c)
\end{split}
\end{equation}

Now we can marginalize each $p(z_n | s_c)$, but instead of using all of $O$, we use only the object in cell $n$:

\begin{equation}
\label{eq:pZcGivenSc}
\begin{split}
    p(z_n|s_c) &=  \sum_i p(z_n | s_c, o_{ni}) \cdot p(o_{ni} | s_c) \\
    & = \sum_i p(z_n | o_{ni}) \cdot p(o_{ni} | s_c) \\
    & =  \sum_i \frac{p(o_{ni} | z_n) \cdot p(z_n)}{p(o_{ni})} \cdot p(o_{ni} | s_c) \\
    & =p(z_n) \cdot \sum_i \frac{p(o_{ni} | z_n)}{p(o_{ni})} \cdot p(o_{ni} | s_c) \\
\end{split}
\end{equation}

We can now make a distinction for $p(o_{ni} | s_c)$ depending on whether $n = c$. The probability for $p(o_{ci} | s_c)$ has a simple interpretation (how likely is the source object to be of a certain category?) and will be provided by the ontology (see section~\ref{sec:ontology}). For the case when $n \neq c$, we can assume that $p(o_{ni} | s_c) \simeq p(o_{ni})$, since knowing there is a gas source in a different point of the environment does not meaningfully modify the probability of this object belonging to any given class. Formally, this can be expressed as follows:

For $s\neq c$, 
\begin{equation}
    p(o_{ni} | s_c) = p(o_{ni} | \neg s_n),
\end{equation}
and the law of total probability states that
\begin{equation}
        p(o_{ni}) = p(s_n) \cdot p(o_{ni} | s_n) + p(\neg s_n) \cdot p(o_{ni} | \neg s_n)
\end{equation}

Since the a priori probability that this specific cell is the source of the gas can always be expected to be orders of magnitude lower than its complement: 
\begin{equation}
        p(\neg s_n) >> p(s_n) \rightarrow   p(o_{ni} | s_c) \simeq p(o_{ni})
\end{equation}

Therefore, continuing from Eq. \ref{eq:pZcGivenSc}, $p (z_n | s_c)$ for $n \neq c$:

\begin{equation}
\begin{split}
    p (z_n | s_c) &= p(z_n) \cdot \sum_i \frac{p(o_{ni} | z_n)}{p(o_{ni})} \cdot p(o_{ni}) \\
    &= p(z_n) \cdot \sum_i p(o_{ni} | z_n) \\
    &= p(z_n)
\end{split}
\end{equation}

And thus, we can show that we can compute $p(s_c|z)$ iterating only through the list of object classes, rather than over the set of all possible maps, drastically reducing the computational complexity from $O(k^n)$ to $O(k)$:

\begin{equation}
\begin{split}
    p(s_c | z) &\propto \prod_{n \setminus c}[p(z_n)] \cdot \sum_i \frac{p(o_{ci} | z_c)}{p(o_{ci})} \cdot p(o_{ci} | s_c) \\
    p(s_c | z) & \propto \sum_i \frac{p(o_{ci} | z_c)}{p(o_{ci})} \cdot p(o_{ci} | s_c)
    \label{eq:prob-semantics-final}
\end{split}
\end{equation}

In this expression, the term $p(o_{ci} | z_c)$ is the current belief of the object classification given all observations, and $p(o_{ci} | s_c)$ is the probability that the gas is being emitted from an object of type $i$, which is defined by the ontology.

\section{Extending the Framework for More Complex Cases}
\label{sec:extensions}
The formulation, as described in this last section, covers the general case. In this section, we will briefly discuss some non-trivial ways in which it can be adapted to the specific requirements of a particular algorithm. 

\subsection{Voxelized Scene Representations}
\label{sec:voxels}
A great number of the state-of-the-art probabilistic GSL algorithms are constrained to 2D estimations due to limitations with their observation models or due to computational constraints. Since the formulation in section~\ref{sec:framework} assumes that the number of source positions considered by the olfactory estimations is equal to the number of object variables, it could be assumed that the semantic representation of the environment must conform to these limitations as well. However, we will now show a simple modification to the calculation of the source probability with semantic information to circumvent this issue and allow for 3D representations of the environment.

The main idea behind this calculation is that a 2D cell can simply be considered a vertical column in 3D space. When the olfaction method gives a probability of the source being inside of one 2D cell, it is the probability that the cell is anywhere in that column. We can apply this same reasoning in the opposite direction, and say that the probability of a 2D cell containing the source given a 3D map of objects is the sum of the probabilities of each of the objects inside of the column that corresponds to that cell:

\begin{equation}
    p(s_c | z) = \sum_{v \in c} p(s_v | z)
\end{equation}

, where $c$ is used to denote a 2D cell in the same way as the previous formulations, and $v$ denotes a 3D voxel. The probability $p(s_v | z)$ can be calculated directly with expression~\ref{eq:prob-semantics-final}.

\subsection{Probabilistic Gas Classification}
\label{sec:gas-classification}
So far we have assumed that we know with certainty the type of gas that is being emitted in the environment. This information is crucial to be able to use the relevant ontology to relate object classes to source probability. It is more realistic to have a formulation that supports cases where there is a probability distribution over a list of possible gas types. This distribution may be updated as the search progresses and the robot receives more measurements.

\begin{figure}
    \centering
    \includegraphics[width=0.7\linewidth]{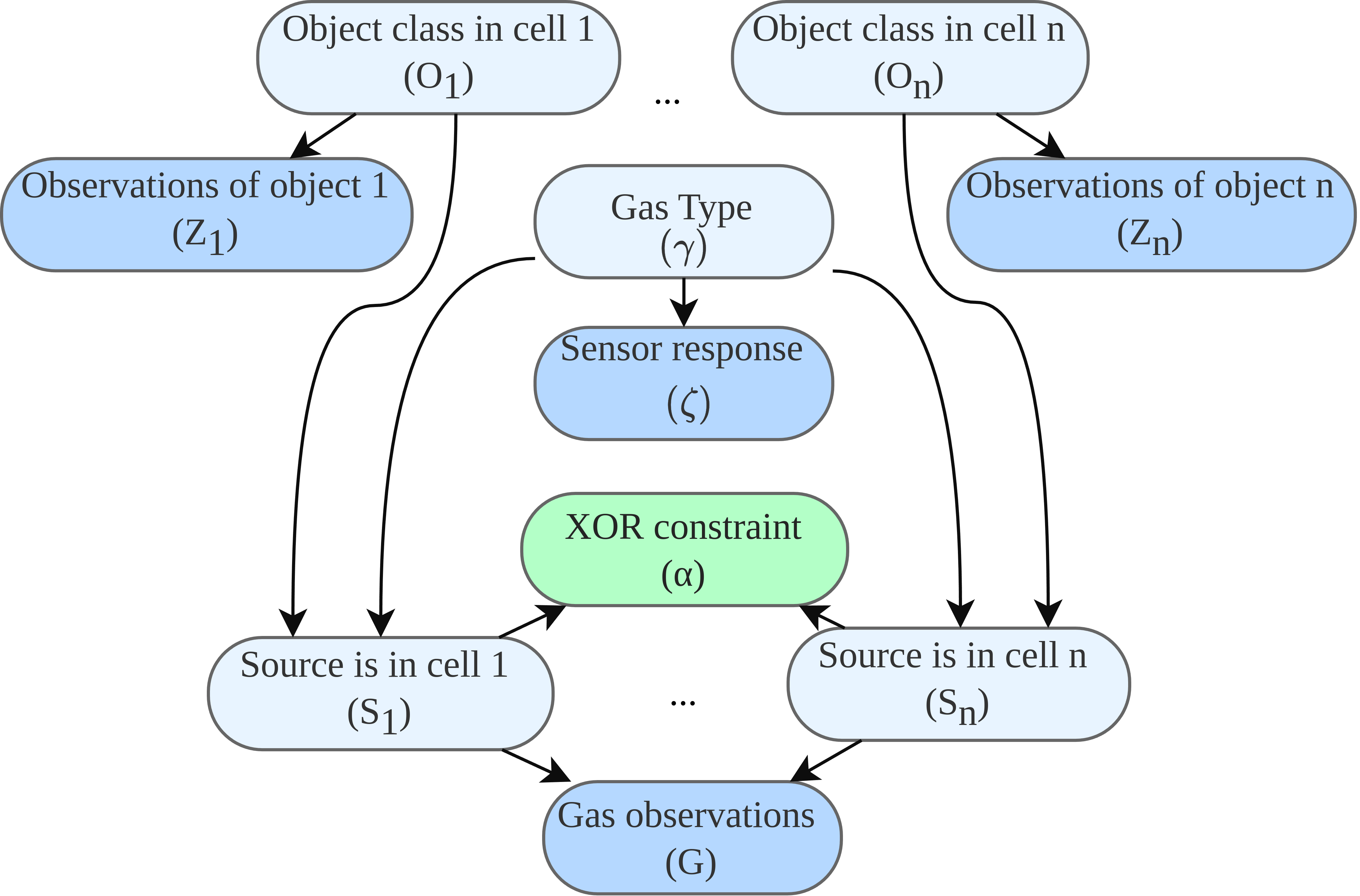}
    \caption{Modified Bayesian network that accounts for dynamic, probabilistic gas classification.}
    \label{fig:bayes-gas-classification}
\end{figure}

This consideration requires modifying the presented formulation slightly. Figure~\ref{fig:bayes-gas-classification} shows the resulting Bayes network after incorporating a categorical random variable $\gamma$ for the type of gas and an observation variable $\zeta$ that denotes the information on which the gas classification is based. 

It is worth noting that we make a distinction between $G$ and $\zeta$, even though both of them relate to olfactory observations. $G$ corresponds to information that is used to reason about the source location: whether there is gas in a certain location or not, how much concentration is detected, the direction of the airflow, \textit{etc.} On the other hand, $\zeta$ refers exclusively to the gas classification. An example of the type of data that $\zeta$ represents is the response curves of an array of different gas sensors when exposed to the same concentration of the gas. This information, lacking the context of where the sample was taken or any other spatial information about the environment, is only relevant to the classification itself.

In this modified model, both knowledge of the object type of a certain cell and knowledge of the gas type being released are necessary to estimate the source location. The ontology to apply in this case would not specify $p(o_c | s_c)$, but rather $p(o_c | s_c, \gamma)$.

The derivation in section~\ref{sec:framework} remains mostly unchanged, and we arrive at a very similar expression:

\begin{equation}
    p(s_c | g, z, \zeta) = p(s_c | g) \cdot p(s_c | z, \zeta)
\end{equation}

The only difference is that the semantics term depends on the gas classification observations $\zeta$. We can expand it with the law of total probability:

\begin{equation}
\begin{split}
    p(s_c | z, \zeta) &= \sum_{\gamma} p(s_c | z, \zeta, \gamma) \cdot p(\gamma | z, \zeta) \\
    & = \sum_{\gamma} p(s_c | z, \gamma) \cdot p(\gamma | z, \zeta)
\end{split}
\end{equation}

Then, applying Bayes' Theorem to the first part:

\begin{equation}
\begin{split}
    p(s_c | z, \zeta) & = \sum_{\gamma} \frac{p(z | s_c, \zeta, \gamma) \cdot p(s_c | \zeta, \gamma)}{p(z| \zeta, \gamma)} \cdot p(\gamma | z, \zeta)
\end{split}
\end{equation}

Now applying it to the denominator:

\begin{equation}
\begin{split}
    p(s_c | z, \zeta) & = \sum_{\gamma} \frac{p(z | s_c, \zeta, \gamma) \cdot p(s_c | \zeta, \gamma) \cdot p(\gamma|\zeta)}{p(\gamma | z, \zeta) \cdot p(z|\zeta)} \cdot p(\gamma | z, \zeta) \\
    & = p(z|\zeta) \cdot \sum_{\gamma} p(z | s_c, \gamma) \cdot p(s_c | \gamma) \cdot p(\gamma|\zeta) \\
    & \propto \sum_{\gamma} p(z | s_c, \gamma) \cdot p(s_c | \gamma) \cdot p(\gamma|\zeta)
\end{split}
\end{equation}

From here, the expansion of the first term is the same as in the case without gas classification, except the final expression has $p(o_{ci} | s_c, \gamma)$ instead of $p(o_{ci} | s_c)$ (which is the value that the ontology gives us). The remaining expression is then:

\begin{equation}
\begin{split}
    p(s_c | z, \zeta) & \propto \sum_{\gamma} p(s_c | \gamma) \cdot p(\gamma|\zeta) \cdot \sum_i \frac{p(o_{ci} | z_c)}{p(o_{ci})} \cdot p(o_{ci} | s_c, \gamma)
\end{split}
\end{equation}

Here, $p(\gamma | \zeta)$ is the current belief for the classification. We can assume that $p(s_c | \gamma) \simeq p(s_c)$ and include it in the normalization term, since knowing the type of gas without any knowledge of the object map does not provide any relevant information about the source location. Then, the final expression reads: 

\begin{equation}
\begin{split}
    p(s_c | z, \zeta) & \propto \sum_{\gamma} p(\gamma|\zeta) \cdot \sum_i \frac{p(o_{ci} | z_c)}{p(o_{ci})} \cdot p(o_{ci} | s_c, \gamma)
\end{split}
\end{equation}

\section{Information Gain Optimization}
A common approach for designing a navigation strategy for GSL algorithms is to consider the expected information gain that would result from each of the considered moves. This idea, sometimes referred to as \textit{infotactic navigation}, or information-gain maximization, was popularized by the Infotaxis algorithm~\cite{vergassolaInfotaxis}, and is also a widely explored concept in other fields of research such as occupancy mapping. 

The main difference when this concept is applied to GSL is that the source location variable itself is not observable, and thus the amount information about it that can be gained from a measurement must be derived from the effect of other, intermediate variables. A metric that can be used to quantify the amount of information about one variable $X$ that is gained by gaining knowledge of a different variable $Y$ is the mutual information $I(X;Y)$, which is defined as the difference between the entropy of $X$, H(X), and the conditional entropy $H(X|Y)$:

\begin{equation}
    I(X;Y) = H(X) - H(X | Y)
\end{equation}

In the case of the formulation presented here, there are two main categories of variables that affect the probability distribution of the source location: the olfactory observations $G$, and the semantic object map $O$. Since these two types of observations are handled separately as per expression~\ref{eq:combined-conditionals}, the information gain $\phi$ obtained at a new measurement pose $p$ can be separated into two terms which correspond to each type of evidence:

\begin{equation}
    \phi(p) = \phi_G(p) + \phi_O(p)
\end{equation}

It is then possible to estimate how much information about the source variable would be gained from the olfactory observations and from the semantic observations in each of the considered measurement poses. Note that we mention poses, rather than just locations, since the semantic observations may include a camera, and thus the facing direction will determine which areas of the map will be observed.

It is also important to note that, since mutual information specifically measures the effect that the measurements will have on the source probability distribution, rather than relying on the uncertainty of the measurements themselves, there is no need to implement weighting parameters or tradeoff mechanisms. The movement strategy can simply rely on the estimated information gain for a given pose, regardless of how much of said information gain is due to the olfactory observations or visual observations.

\begin{figure}
    \centering
    \includegraphics[width=1\linewidth]{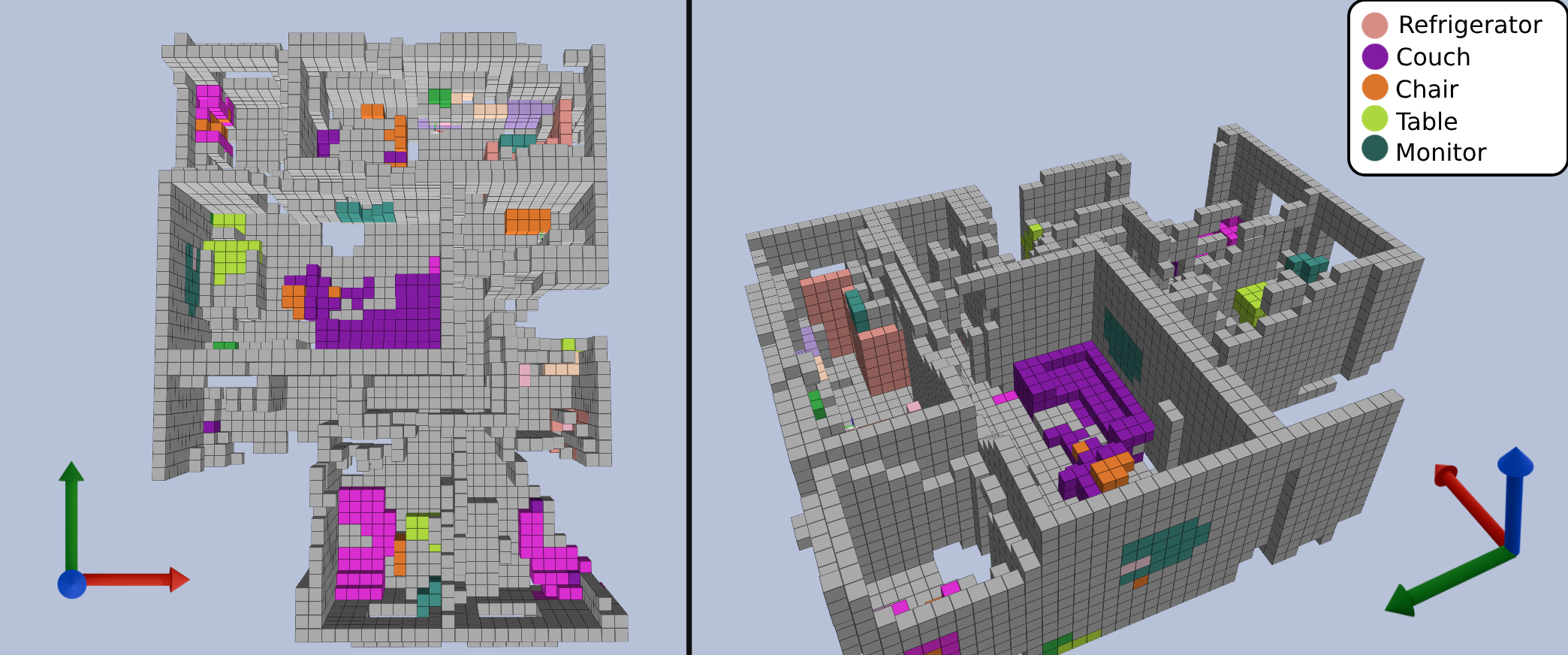}
    \caption{Voxelized semantic map reconstructed with Voxeland~\cite{voxeland} during the search process. The colored voxels denote a specific semantic category.}
    \label{fig:voxeland}
\end{figure}

We will not discuss here the subject of calculating the mutual information of the source and the olfactory observations, since our formulation does not specify the details of the olfactory observation model. For the semantic information, however, let us consider the case where the robot is equipped with a camera that it can use to detect new objects. The information gain about $S$ from knowing the value of $O_n$ is:

\begin{equation}
    I(S; O_n) = H(S) - \sum_{i} p(o_{ni}| z) \cdot H(S| o_{ni})
\end{equation}

We can then define the set $F_p$ of all the cells that fall inside of the FOV of the camera from pose $p$, and calculate the semantic information gain as:

\begin{equation}
    \phi_O(p) = \sum_{n \in F_p} I(S; O_n)
\end{equation}

\section{Validation}
In this section we present some experimental results obtained by extending PMFS~\cite{OjedaPMFS} an olfaction-based GSL algorithm, with semantic information. 

The experiments are carried out in simulation, using Gaden~\cite{monroyGADEN3DGas2017} as a gas dispersion simulator. We leverage its integration with the Unity Engine~\cite{ojeda_2021_sensors} to handle the visual part of the simulation, using an environment from the VGR Dataset~\cite{vgr_mro_dataset} to provide a complex environment that features many object of multiple types.

To handle the dense semantic map, we rely on Voxeland~\cite{voxeland}, a voxel-based framework that handles semantic mapping in a probabilistic manner and can provide a probability distribution over the list of object classes for each voxel. In order to integrate this voxel map into our formulation, we use the method described in section~\ref{sec:voxels}. Figure~\ref{fig:voxeland} shows the semantic map of the experimental scenario that is reconstructed on-line by Voxeland during the search process. We use Mask-RCNN~\cite{he2018maskrcnn} to detect and classify objects in the images captured by the camera.

\begin{figure}
    \centering
    \includegraphics[width=0.7\linewidth]{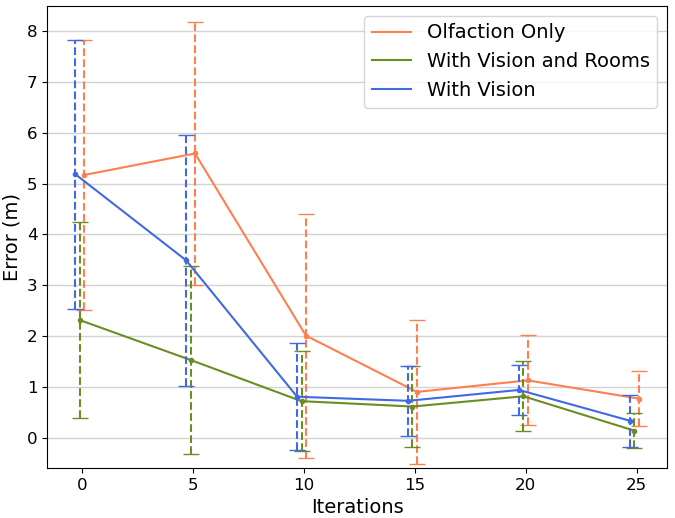}
    \caption{Evolution of the error in the estimated source position over the course of the search operating with different amounts of semantic information. It can be observed that the version with no semantic data is slower to identify the correct area, and achieves lower precision in the final declaration.}
    \label{fig:error-chart}
\end{figure}

Our main interest here is to analyze the effect of adding semantic data to the source localization process. Therefore, we compare a version of the PMFS algorithm running by itself that relies only on olfactory data to a version that has semantic information. With regard to the semantic information itself, we consider two cases. In the first case, the only information about the object map that is available to the robot is the scene reconstruction that is generated from visual observations. In the second case, a simple 2D map that marks the room boundaries and their type is also available. This information about the rooms is treated as an additional observation, and serves to modify the probability distribution over the object classes for each cell (\textit{e.g.} a cell that is in a room labeled \textit{kitchen} has a higher than normal probability of being an oven). For the run shown here, the substance being emitted is smoke, and the source is an oven.

\begin{figure}
    \centering
    \includegraphics[width=0.6\linewidth]{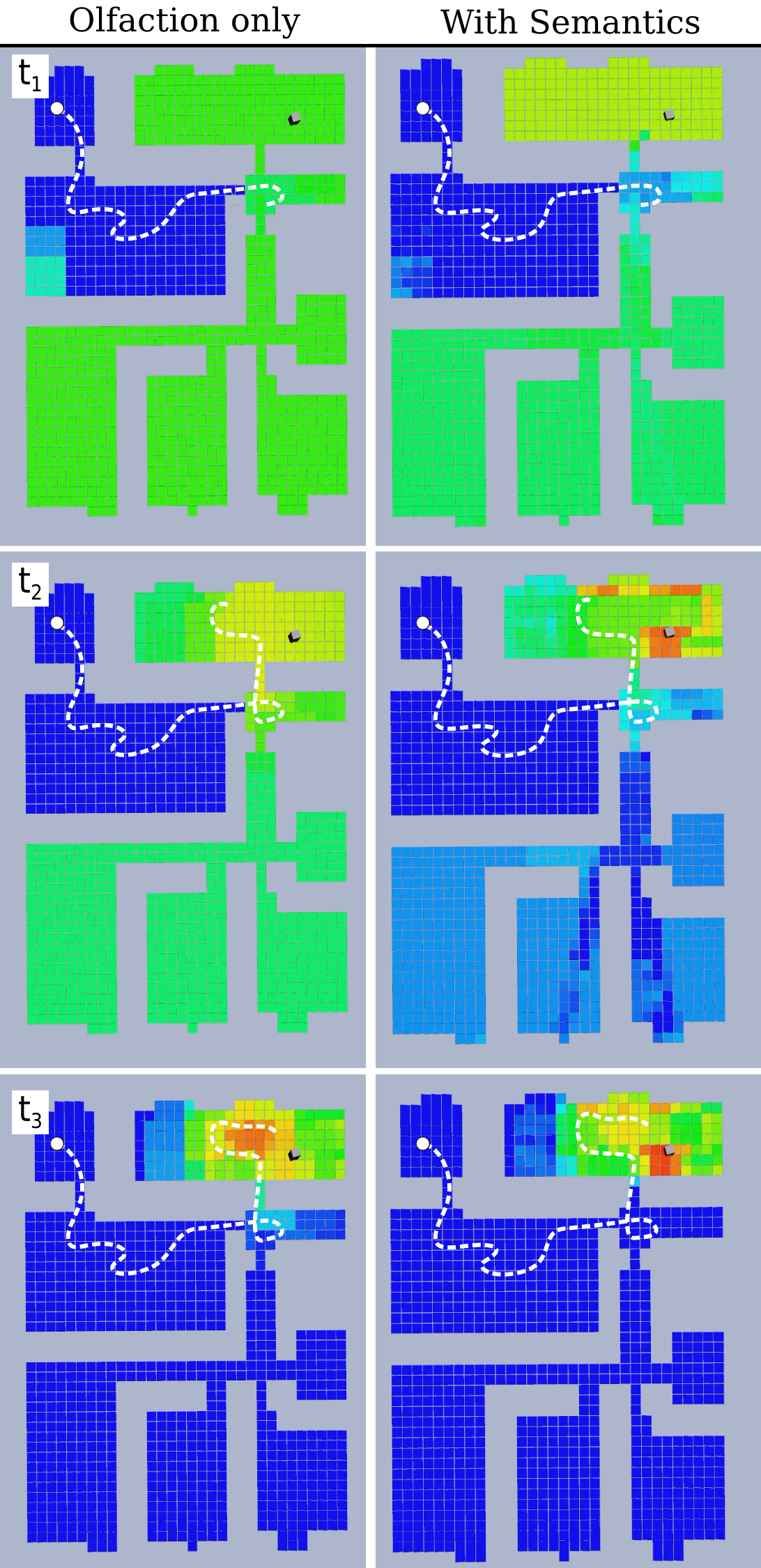}
    \caption{Progression of the source probability distribution through the search process. It can be observed that with semantics (room categories and camera observations) the localization converges more quickly to the correct position and achieves higher precision on the source declaration. The dashed line indicates the path followed by the robot.}
    \label{fig:progression}
\end{figure}

We focus here on the precision of the produced estimation of the source location. To make this comparison fair, all three methods follow the same path and obtain the same observations. Figure~\ref{fig:error-chart} shows the results of this comparison with the three versions of the method. The error of the source estimation is reported as the distance between the expected value of the probability distribution and the ground-truth source location, and the error bars show the standard deviation of the distribution. It can be observed that the versions that include semantic information achieve higher precision in the source declaration, as recognizing the source object as an oven causes the probability distribution to favor those cells as a likely source location. Figure~\ref{fig:progression} shows the progression of the source probability distribution during the search.

\begin{figure}
    \centering
    \includegraphics[width=0.8\linewidth]{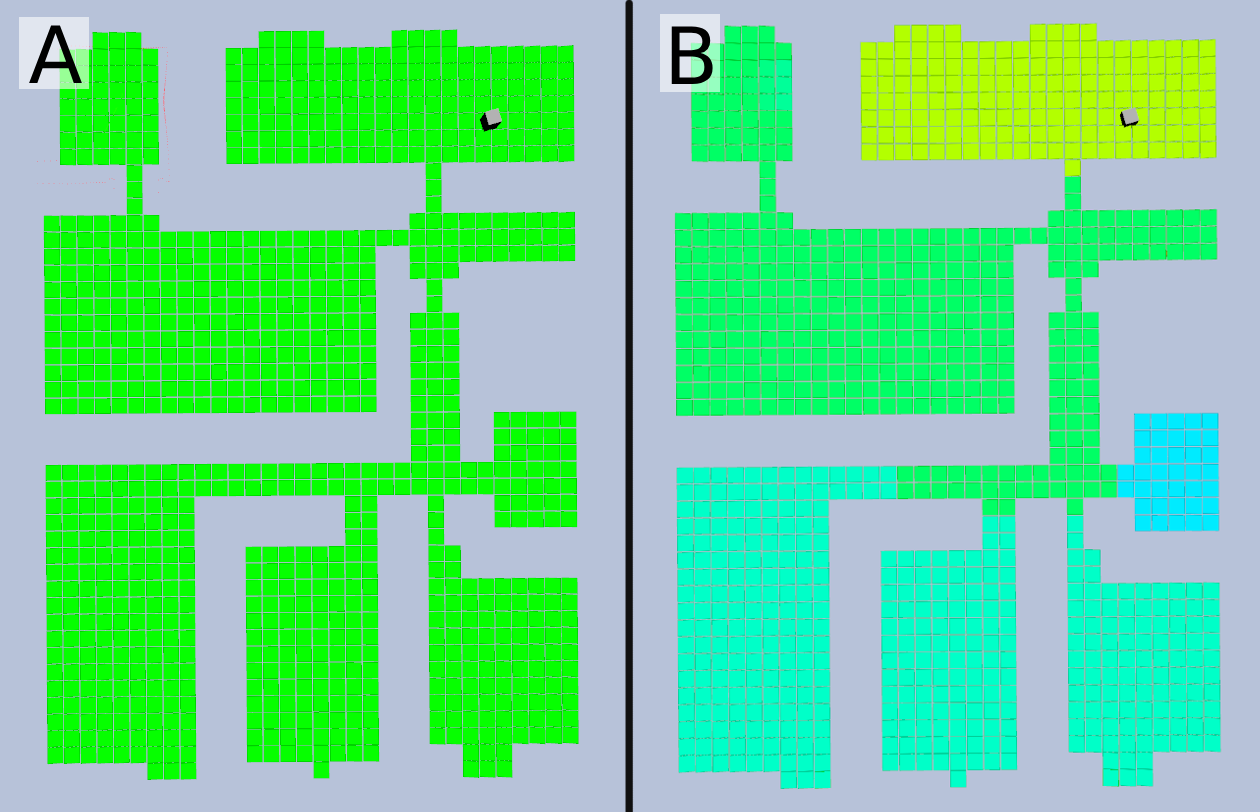}
    \caption{Comparison of the source probability distributions at the start of the search, (A) with no previous information, (B) with a map of the room categories.}
    \label{fig:room-types}
\end{figure}

It can also be observed in Figure~\ref{fig:error-chart} that the inclusion of the room categories greatly reduces the error during the initial part of the search, even though the final result is not significantly better when compared to the other semantic version. This effect is caused by the cells in the vicinity of the source having a higher estimated probability of being of a relevant object class before being observed, due to being in a kitchen (see Figure~\ref{fig:room-types}). This could be used to guide the movement of the robot during this initial phase directly to the areas of highest interest, making the algorithm faster at finding the source. The fact that both semantic methods achieve the same precision in the end is due to the fact that, given enough observations of a certain object, the visual data dominates the estimation of its object class.

\section{Future Work}
This paper has presented a probabilistic formulation for integrating semantic information into the problem of gas source localization. We formally described the relevant concepts, and derived the equations used to estimate the source location given this new data.

Future work must include a more thorough experimental evaluation of the results obtained with this technique. Given the abstract nature of the formulation presented here, the semantic information should be combined with different olfactory GSL methods to analyze its impact depending on the characteristics of said methods. 

A more in-depth analysis of the subject of movement strategies and information gain optimization is also desirable, as the discussion is not exhaustive. 

Extensions to the formulation, such as modifications for working with sparse semantic information, are also a possible avenue for future research.


\bibliographystyle{plain}
\bibliography{main.bib}
\end{document}